\pdfoutput=1

\documentclass[11pt]{article}

\usepackage[preprint]{acl}

\usepackage{times}
\usepackage{latexsym}
\usepackage{xcolor}
\usepackage{amsthm}   

\theoremstyle{plain}                  

\usepackage[T1]{fontenc}

\usepackage[utf8]{inputenc}

\usepackage{microtype}

\usepackage{inconsolata}

\usepackage{graphicx}
\usepackage{booktabs}
\usepackage{multirow}   
\usepackage{amsmath}
\usepackage{caption}
\usepackage{float}

\title{Uncertainty-Aware Answer Selection for Improved Reasoning in Multi-LLM Systems}

\author{
 \textbf{Aakriti Agrawal \textsuperscript{1}},
 \textbf{Rohith Aralikatti \textsuperscript{2}},
 \textbf{Anirudh Satheesh \textsuperscript{1}},
 \textbf{Souradip Chakraborty \textsuperscript{1}},
\\
 \textbf{ Amrit Singh Bedi \textsuperscript{3}},
 \textbf{Furong Huang\textsuperscript{1,4}}
\\
 \textsuperscript{1} University of Maryland,
 \textsuperscript{2} Hilabs,
 \textsuperscript{3} University of Central Florida,
 \textsuperscript{4} Capital One
\\
 \small{
   \textbf{Correspondence:} \href{mailto:agrawal5@umd.edu}{agrawal5@umd.edu}
 }
}

\begin{document}
\maketitle
\begin{abstract}
Large Language Models (LLMs) have demonstrated exceptional capabilities, yet selecting the most reliable response from multiple LLMs remains a challenge, particularly in resource-constrained settings. Existing approaches often depend on costly external verifiers, human evaluators, or self-consistency techniques that require multiple samples from a single model. While multi-LLM systems produce more diverse responses than single models and thus have greater potential, they often underperform compared to single LLM self-consistency. We propose a principled, novel and computationally efficient method to select the best response from multiple different LLMs using a calibrated log-likelihood score, implicitly leveraging the inherent knowledge and confidence of these models. Our method demonstrates improvements of approx. 4\%, 3\%, and 5\% across both debate (multi-round LLM discussions) and non-debate (Best-of-N with multiple LLMs) settings on GSM8K, MMLU (6 subsets), and ARC datasets respectively\footnote{Code: https://github.com/Aakriti05/multi-llm-uncertainty}.


\end{abstract}

\section{Introduction}
Large language models (LLMs) have achieved remarkable advancements, with state-of-the-art (SOTA) models now approaching or even surpassing human performance. With rapid advancements, an array of pre-trained LLMs and LLM APIs have become readily accessible, each exhibiting distinct strengths across specialized tasks. For instance, models like GPT-deep research excel in analytical and research-oriented tasks, whereas GPT-o3 is tailored towards programming and coding. Similarly, there are domain-specific expert models optimized for tasks in physics, algebra, and other fields.

This proliferation raises an important question: \textbf{How can we optimally leverage the diverse capabilities of multiple specialized LLMs to produce the most accurate and reliable answers without incurring additional training costs} (given the substantial resource demands associated with training) ? This challenge becomes especially critical when high-quality external judges or evaluators—capable of assessing answers across multiple specialized domains—are unavailable, expensive, or even infeasible due to the superhuman capabilities of modern LLMs \cite{burns2023weak, agrawal2024ensemw2s}.

Existing approaches have tried exploring this but \textbf{(Challenge 1)} suffer from reliance on external information such as: (1) \textit{external verification models} \cite{xi2024enhancing}, (2) \textit{human or LLM-based judges} \cite{chan2023chateval, khan2024debating, li2024llms}, or (3) \textit{reward models trained explicitly for response ranking}. These methods involve significant resource overhead and are infeasible when existing models reach superhuman performance.

Additional major challenge is that existing methods from single-LLM literature like majority voting with self-consistency \cite{wang2023selfconsistency}, self-reflection \cite{renze2024self}, or metric-based selection \cite{kang2025scalable} (e.g., perplexity, self-certainty) \textbf{(Challenge 2)}  demand extensive sampling or \textbf{(Challenge 3)} are infeasible for multi-LLM contexts due to inherent differences in outputs from a multi-LLM system.
\textbf{(Challenge 4)} A simplistic multi-LLM adaptation of self-consistency, which selects the most frequently generated answer, does not effectively exploit inter-model reasoning and hence misses potential performance gains \cite{du2023improving}.


To address the above challenges, in this paper, we propose a novel and computationally efficient method method to aggregate responses from diverse LLMs and systematically select the best answer without relying on external verifiers (Challenge 1), without requiring extensive sampling (Challenge 2), and effectively use diverse multi-LLMs (Challenge 3, 4).
Specifically, we introduce \textbf{uncertainty estimation-based answer selection from multi-LLM systems}, which employs a calibrated log-likelihood-based selection metric that implicitly leverages the inherent knowledge and confidence of the given models, further improving response accuracy and reducing computational overhead. Our method is built on the hypothesis that a model (or expert) trained on a specific example will exhibit high confidence (i.e., low uncertainty), while models unfamiliar with the example will show higher uncertainty. Assuming the training data is correct, the most confident expert is likely to provide the correct answer and our aim to find that expert. 

Thus, the primary contributions are:

\begin{enumerate}
\setlength{\itemsep}{0pt}
    \item We propose a principled calibration technique for aligning log-likelihoood based uncertainty scores across different models to select the best answer. This is because directly comparing per-token likelihoods across models is theoretically flawed, as each model has distinct parameters and logit distributions.
    \item Our approach is also computationally efficient: by using teacher-forcing, we compute calibrated scores with a single forward pass, avoiding costly autoregressive decoding. It is light weight and simple as it does not require external verifiers, reward models, human or LLM judges.
    \item Our method demonstrates strong empirical performance across both debate (multi-round multi-LLM discussions) and non-debate (Best-of-N with multiple LLMs) settings on GSM8K, MMLU (6 subsets), and ARC datasets showing improvement of 4\%, 3\% and 5\% respectively. It \textbf{achieves greater improvements in multi-LLM settings compared to single LLM}, further reinforcing the potential of a diverse-answer-generating multi-LLM system to boost overall performance.
    \item We also show comparison with other metrics and theoretical justification for our calibrated metric.
\end{enumerate}




\begin{figure*}[t]
    \centering
    \includegraphics[page=2, trim=0cm 0.5cm 0.5cm 0cm, clip, width=\textwidth]{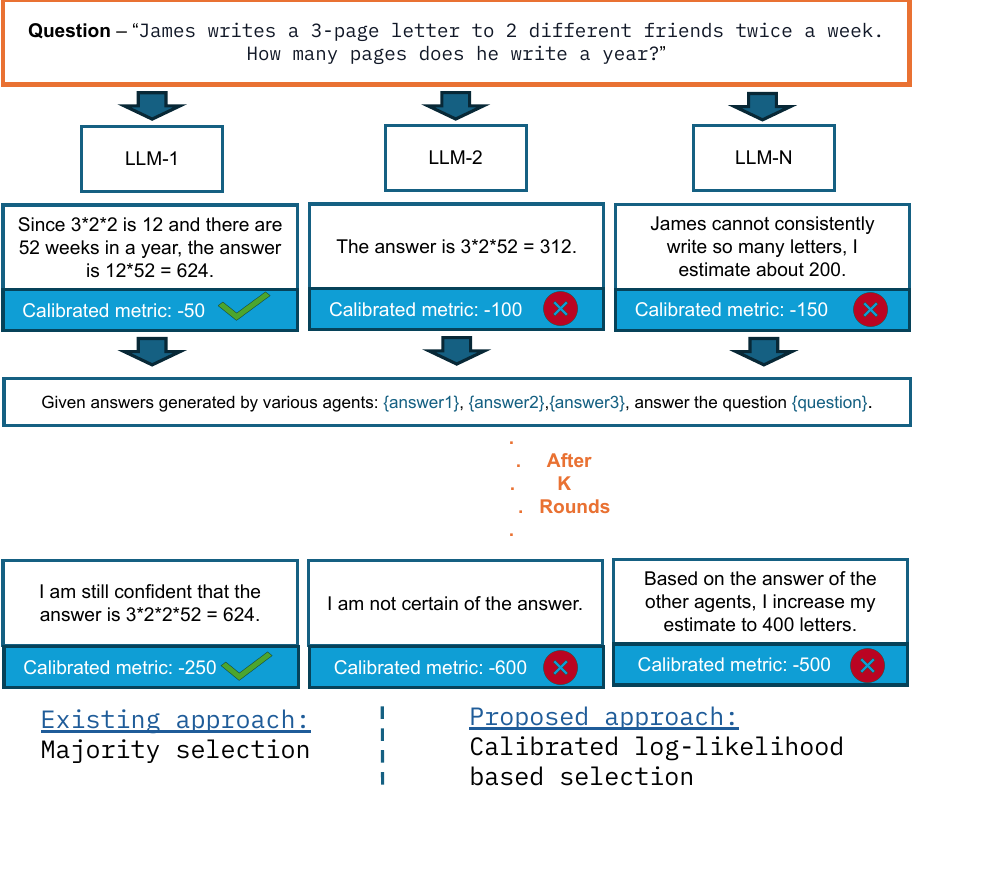}
    \vspace{-15mm}
    \caption{
        Diagram illustrating our proposed calibrated log-likelihood based metric for answer selection in a multi-LLM system. The multi-LLM system shows a debate setting which goes on for $K$ rounds. When $K=1$, it is best-of-N setting when answers are independently sampled from each LLM and best answer is selected without exposing the answer to the other LLMs. The best answer is chosen based on calibrated log-likelihood scoring. Our-approach outperforms random selection when there is no clear majority answer.
    }
    \label{fig:multi_llm_debate}
\end{figure*}

\vspace{-1.5mm}
\section{Uncertainty-Aware Answer Selection}

We now describe our approach for aggregating and selecting optimal answers in a multi-LLM system. While methods for optimal answer selection—commonly used in best-of-N strategy—are well-studied in single-LLM contexts \cite{kang2025scalable}, directly extending them to multi-LLM environments poses significant challenges. Specifically, diverse LLMs (1) produce outputs varying significantly in format and length, (2) differ widely in their model weights and architectures, and (3) yield uncertainty scores that are typically incomparable across models due to a lack of calibration.

To address these issues, we propose a \textbf{calibrated log-likelihood metric}, which integrates smoothly with the interactive multi-LLM debate as well non-interactive multi-LLM best-of-N setup. We first detail our proposed metric and then provide theoretical insights and analysis to illustrate its enhanced performance. We also provide discussion on its computational efficiency. Figure~\ref{fig:multi_llm_debate} illustrates how our calibrated selection approach surpasses traditional majority-voting strategies for a multi-LLM setup. (Note: A non-debate best-of-N setting is equivalent to first round of debate setting.).

\subsection{Calibrated Log-Likelihood Metric}

Consider a set of \(N\) LLMs \(\{\pi_1, \pi_2, \ldots, \pi_N\}\) each generating a response to a given prompt \(x\). We denote the response from the $i^{th}$ model \(\pi_i\) as \(Y_i\).  In the first debate round, each LLM independently produces a response \(Y^1_i = \{y^{1}_1, y^{1}_2, \ldots, y^{1}_T\}\), where \(T\) is the number of tokens with log-likelihood computed as:
\vspace{-4mm}
\[
\log P_{\pi_{i}}\big(Y^1_i \mid x\big) = \frac{1}{T} \sum_{t=1}^{T} \log P_{\pi_i}\big(y^{i}_t \mid x, y^{i}_{<t}\big),
\]
\vspace{-0mm}
where \(P_{\pi_i}\big(y^{i}_t \mid x, y^{i}_{<t}\big)\) denotes the conditional probability of token \(y^{i}_t\) given the prompt \(x\) and preceding tokens \(y^{i}_{<t}\) for model \(\pi_i\).

In subsequent rounds (\(K > 1\)), model $\pi_i$'s response $Y^K_i$ depends on both the prompt and previous responses. Denoting previous responses as $X = \{Y^{k}_{j}\}_{\substack{\forall j\in[1,N],  \forall k < K}}$, the log-likelihood is then computed as, 

\vspace{-3mm}


\[
\scalebox{0.9}{$
\log P_{\pi_i}\!\left(Y_i^K \mid x, X \right) 
= \frac{1}{T} \sum_{t=1}^T \log P_{\pi_i}\!\left(y_{i,t}^K \mid x, X, y_{<t}\right)
$}
\]

where,
$Y_i^K = \left( y_{i,1}^K,\, y_{i,2}^K,\, \ldots,\, y_{i,T}^K \right)$

To select the best final response from $\{Y^{K}_{i}\}_{\substack{\forall i\in[1,N]}}$, we define a calibrated log-likelihood-based scoring mechanism as:
\vspace{-3mm}
\begin{equation}
Score (Y_{j}) = \frac{1}{N}\sum_{i=1}^{N}\log P_{\pi_i}(Y^K_j \mid x, X)
\label{eq:calibration}
\end{equation}

This metric measures consensus among models by averaging response likelihood across all models, providing a calibrated measure of answer quality.

\paragraph{Our Calibrated Metric with any Uncertainty Method:} Given a response \(C\) to a prompt \(p\) from LLM \(i\) with uncertainty metric \(M_i(C \mid p)\), we define the calibrated metric \(M_c\) in a multi-LLM setting as:
\vspace{-3mm}

\[
M_c(C \mid p) = \frac{1}{N} \sum_{i=1}^N M_i(C \mid p), \]
where $p = x \quad \text{or} \quad p = (x, Y_i^K)$


While uncalibrated scores may work in some empirical settings, they can introduce \textbf{bias toward the originating model and compromise fairness or consistency}. Therefore, this calibration step is crucial and represents a important contribution of this work. To motivate the general use of this diverse calibration, we also compare the performance of log-likelihood with other metrics such as Gini impurity, self-certainty \cite{kang2025scalable} and entropy in Table \ref{tab:gsm_arc} and \ref{tab:mmlu}. Please refer to the appendix \ref{appx:metric} for more details. Our results show that log-likelihood tends to perform better or similar to other metrics.

Note: In practice, we apply our calibrated log-likelihood scoring method only if a clear majority doesn't emerge. This is because models are likely to converge to a common answer only if its correct (assuming they are trained on correct answers). The likelihood of all models generating the same incorrect answer is very low. Thus, applying model uncertainty metric when there is consensus is redundant and computationally inefficiency. In appendix \ref{appx:tie_break} we provide results for apply calibrated log-likelihood to all cases vs only tie-break cases. As expected, the gain beyond tie-break only is marginal or none, confirming that the extra computation (for non-tie break cases) is unnecessary.


\begin{table*}[t]
\resizebox{\textwidth}{!}{%
\begin{tabular}{l|llllll}
\hline
                           & \multicolumn{6}{l}{GSM8K}                                                                                                      \\ \hline
                           & Random                    & Log-Likelihood            & Self-Certainty & Gini Score & Entropy & Upper Bound \\ \hline
Best-of-N (Q=9)             & \multicolumn{1}{r}{82.55}                     & \multicolumn{1}{r}{82.55}                     &       \multicolumn{1}{r}{-}            &       \multicolumn{1}{r}{-}              &     \multicolumn{1}{r}{-}           &     \multicolumn{1}{r}{92.65}       \\
Best-of-N (Q=1 | L=1 | M=1) &   \multicolumn{1}{r}{47.87}                         &       \multicolumn{1}{r}{\textbf{70}}                  &            \multicolumn{1}{r}{-}         &               \multicolumn{1}{r}{-}      &     \multicolumn{1}{r}{-}           &         \multicolumn{1}{r}{70}    \\
Best-of-N (Q=3 | L=3 | M=3) &   \multicolumn{1}{r}{76.18}                         &       \multicolumn{1}{r}{\textbf{77.16}}                  &            \multicolumn{1}{r}{-}         &               \multicolumn{1}{r}{-}      &     \multicolumn{1}{r}{-}           &         \multicolumn{1}{r}{89.51}    \\
Debate (Q | L | M) & \multicolumn{1}{r}{81$\pm$0.24}                        & \multicolumn{1}{r}{\textbf{84.88}} & \multicolumn{1}{r}{84.88} & \multicolumn{1}{r}{84.57} & \multicolumn{1}{r}{84.34} & \multicolumn{1}{r}{90.00}  \\ \hline
                           & \multicolumn{6}{l}{ARC}                                                                                                        \\ \hline
Best-of-N (Q=3)             &    \multicolumn{1}{r}{85.88}                       &         \multicolumn{1}{r}{\textbf{85.97}}                  &        \multicolumn{1}{r}{-}             &                \multicolumn{1}{r}{-}      &          \multicolumn{1}{r}{-}      &         \multicolumn{1}{r}{-}    \\
Best-of-N (Q=1 | L=1 | M=1) & \multicolumn{1}{r}{83.90$\pm$0.01} & \multicolumn{1}{r}{\textbf{89.00}}    & \multicolumn{1}{r}{88.91} & \multicolumn{1}{r}{89.00}    & \multicolumn{1}{r}{89.00}    &  \multicolumn{1}{r}{94.62}  \\
\hline
\end{tabular}%
}
\caption{Performance comparison demonstrating the effectiveness of calibrated metrics for tie-breaking. For GSM8K, we assess three scenarios: (a) best-of-N with the best-performing single model (Qwen2.5-7B-Instruct) using $N=9$, (b) best-of-N with samples pooled evenly from all three models ($N=3$ per model), and (c) a three-model debate setting similar to \cite{du2023improving} with three rounds and one sample per round per model. In all scenarios, the number of LLM calls remains identical. For the ARC dataset, we report results for single-LLM best-of-N ($N=3$) and multi-LLM best-of-N ($N=1$ per model).}
\label{tab:gsm_arc}
\end{table*}

\subsection{Why Does Calibrated Log-Likelihood Improve Multi-LLM Performance?} 
Diversity has shown to be the major reason of improving best-of-N performance \cite{wang2025diversified} in single LLM systems. The upper-bound results in Table~\ref{tab:gsm_arc} show the diversity offered by multiple LLMs showing significant performance potential. Our calibrated log-likelihood metric exploits this intrinsic diversity to find the best answer to enhance overall system accuracy.

According to \cite{yadkori2024believe}, models demonstrate higher confidence and reduced hallucinations when familiar with a given data point or scenario. Importantly, confident models maintain their confidence even when prompted with incorrect answers, unlike less confident models whose uncertainty notably increases. Consequently, model confidence acts as a strong proxy for response correctness. Our proposed uncertainty-aware metric directly captures this relationship, allowing effective discrimination between correct and incorrect responses.

Direct application of standard uncertainty metrics \cite{kang2025scalable} is unsuitable for multi-LLM because of different $\pi_i$. To address this, our method normalizes log-likelihood scores across multiple models, as shown in Equation~\ref{eq:calibration}, making scores directly comparable. Additionally, we normalize log-likelihood values by sentence length to account for variability in response lengths across models. A theoretical justification is provided in Appendix~\ref{appx:theory}. To validate our approach, we empirically demonstrate (see Fig. \ref{fig:ll_histogram} in the Appendix) that there is a concentration of correct answers at lower values of log-likelihood. This observation reinforces our use of uncertainty-based metrics for robust multi-LLM answer selection.

\subsection{Computational Cost of the Proposed Approach.} 
Our calibrated scoring is computationally efficient as it does not require any additional decoding; it repurposes the already‑generated completions and merely queries each model once per completion. To explain this further lets denote the number of language models in the ensemble as $N$ and the (average) number of tokens in a completion as $L$. With one completion per model, the \textit{calibrated log‑likelihood} procedure evaluates every (model, completion)  pair exactly once, thus making a total of $N^{2}$ \textit{teacher‑forced} forward passes:

\[
Cost_{\text{cal}} 
= N^2 \times 
\underbrace{\bigl(1 \text{ forward pass over } L\bigr)}_{\text{parallel over time}}
= O(N^2)
\]

The total time is proportional to only $N^{2}$ as \textit{all} $L$ tokens are processed in parallel on modern accelerators. If we compare our method with N generations per model, the expected cost of the procedure will be proportional to $L$ as well because of autoregressive generation. For long, reasoning‑heavy answers $(L >> N)$—the increase in computation is thus significant. 

\[
Cost_{\text{gen}} = O(N^2L)
\]

Here $Cost_{\text{cal}}$ and $Cost_{\text{gen}}$ refer to the number of forward passes after GPU parallelization. Hence tie‑breaking via extra generation is roughly a factor of $L$ more expensive than calibrated scoring.

\section{Experimental Setup and Results}

\begin{table}[htbp]
\resizebox{\columnwidth}{!}{%
\begin{tabular}{lrrrrrrr}
\hline
 &
  \multicolumn{1}{l}{FL} &
  \multicolumn{1}{l}{HSM} &
  \multicolumn{1}{l}{EM} &
  \multicolumn{1}{l}{CM} &
  \multicolumn{1}{l}{PHI} &
  \multicolumn{1}{l}{AA} &
  \multicolumn{1}{l}{Avg} \\ \hline
Random  & 47.20           & 46.38          & 79.01          & 40.81          & 69.77          & \textbf{43.87}          & 54.51          \\
Ours (Q=1 | L=1 | M=1)     & \textbf{49.60}  & \textbf{50.57} & \textbf{82.23} & \textbf{43.87} & \textbf{70.09} & {42.85} & \textbf{57.92} \\
Upper Bound  & 72.00           & 65.90         & 93.00         & 64.00          & 81.00         & 57.00          & 72.15          \\

\hline
Random & 50.79          & 48.89          & 79.36          & 37.00             & 69.45          & 48.00             & 56.54          \\
Ours Debate (Q | L | M)   & \textbf{53.17} & \textbf{51.85} & \textbf{82.8}  & \textbf{37.00}    & \textbf{72.26} & \textbf{51.00}    & \textbf{58.98} \\ 
Upper Bound  & 73.80           & 70.70          & 93.90         & 65.00         & 83.90         & 68.00          & 75.88         \\
\hline
\end{tabular}%
}
\caption{Accuracy comparison on MMLU dataset subsets: Formal Logic (FL), High School Math (HSM), Elementary Math (EM), College Mathematics (CM), Philosophy (PHI), and Abstract Algebra (AA). We evaluate random tie-breaking versus proposed metric-based tie-breaking in two settings: (a) best-of-N sampling across three models, and (b) three-model debate. Our approach consistently outperforms the baseline.}
\label{tab:mmlu}
\end{table}

\textbf{Baselines.} We evaluate our proposed metric for multi-LLM setups by comparing it against commonly used majority voting with random tie-breaking \cite{du2023improving}, considering both interactive multi-LLM debate and non-interactive best-of-N settings. We utilize \textbf{three models:} Qwen2.5-7B-Instruct, Ministral-8B-Instruct-2410, and Llama-3.1-8B-Instruct. The \textbf{evaluation datasets} are GSM8K \cite{cobbe2021training}, MMLU \cite{hendrycks2020measuring} (comprising 8 subsets), and ARC \cite{allenai:arc}. Further experimental details are provided in Appendix \ref{appx:experimental_setup}.

Additionally, we compare against single-LLM best-of-N sampling using the best-performing model (Qwen2.5-7B-Instruct). This comparison allows us to assess the relative advantage provided by multi-LLM systems when using our proposed approach. To ensure fairness, we maintain an equal number of total LLM calls across all evaluated scenarios.

\subsection{Tie-breaking Using Calibrated Metrics Outperforms Random Tie-breaking}

Our experiments in table \ref{tab:gsm_arc}, \ref{tab:mmlu} demonstrate that employing calibrated metrics for tie-breaking yields significant performance improvements compared to random tie-breaking. Specifically, for GSM8K and MMLU, we observe absolute accuracy improvements of \textbf{3.88\% and 2.44\%} respectively, within the multi-LLM debate setting. For best-of-N sampling, we observe absolute accuracy improvements of \textbf{1\%, 3.41\% and 4.9\%}, respectively for GSM8K, MMLU and ARC datasets. Furthermore, this approach allows the multi-LLM setting to outperform single-model best-of-N baselines using Qwen2.5-7B-Instruct as seen in table \ref{tab:gsm_arc}, even when the total number of LLM calls remains constant.

\subsection{All Calibrated Metrics Provide Similar Performance}

Table \ref{tab:gsm_arc} compares various calibrated metrics for tie-breaking across two datasets (GSM8K and ARC). These calibrated metrics are well-established alternatives to majority voting in single-LLM best-of-N scenarios, as previously reported \cite{kang2025scalable}. Our findings indicate that performance differences among the calibrated metrics are comparable, but as seen from Table \ref{tab:gsm_arc} calibrated log-likelihood performs the best.

\section{Conclusion}
\vspace{-1mm}
We introduced a calibrated log-likelihood-based selection framework to enhance multi-LLM systems. By leveraging uncertainty estimation, our method selects the most confident response, reducing reliance on costly external verifiers and extensive sampling. Our approach outperforms random selection and majority voting with same model calls, making it a cost-effective solution. Additionally, we highlight the benefits of diverse model reasoning in multi-LLM debate. Future work can explore adaptive sampling and extend our method to broader reasoning tasks.
\section{Limitation}

Our method is primarily effective in low-cost settings, where the number of LLM calls is limited. In high-cost settings—where a large number of responses can be generated—the likelihood of a non-majority-voted answer decreases. As a result, the effectiveness of our log-likelihood-based selection in improving performance over random selection diminishes significantly.

Additionally, our approach is specifically designed for multi-LLM settings, where diverse models generate a broader range of responses. This diversity encourages deeper reasoning and exploration of alternative answers, increasing the likelihood of finding and converging on the correct solution. In such scenarios, our selection method is particularly valuable in identifying the most confident response among the generated outputs. However, in single-LLM self-consistency settings, where the responses are inherently less varied, our method may provide limited benefits.

\newpage
\section{Acknowledgments}

Agrawal, Satheesh, Chakraborty and Huang are supported by DARPA Transfer from Imprecise and Abstract Models to Autonomous Technologies (TIAMAT) 80321, DARPA HR001124S0029-AIQ-FP-019, DOD-AFOSR-Air Force Office of Scientific Research under award number FA9550-23-1-0048, National Science Foundation NSF-IIS-2147276 FAI, National Science Foundation NAIRR240045, National Science Foundation TRAILS Institute (2229885). Private support was provided by Peraton.

\bibliography{custom}
\newpage
\appendix
\section{Related Works}

\paragraph{Multi-LLM Systems:} 

Recent work on multi-LLM systems has explored various strategies to enhance performance and evaluation. Mixture-of-Experts (MoE) models, such as Uni-MoE, integrate multiple specialized components within a unified architecture \cite{li2025uni}. EnsemW2S \cite{agrawal2024ensemw2s} combines diverse LLM's token-level probabilities using Adaboost inspired weighing mechanism to improve generalization on complex reasoning tasks. Other methods include PromptEval, which estimates performance across prompts for robust evaluation \cite{polo2024efficient}, LLM as judge \cite{chan2023chateval, khan2024debating, li2024llms} and debate/discussion among agents \cite{du2023improving}. Additionally, research on MoE routing weights suggests their potential as complementary embedding models \cite{li2024your}. These works highlight the growing interest in optimizing multi-LLM collaboration for improved reasoning and evaluation.

\paragraph{Single LLM Self-Improvement:}

To improve reasoning and mitigate inconsistencies in LLM outputs, recent work has explored feedback-based learning. Self-consistency \cite{wang2023selfconsistency} aggregates multiple sampled outputs via majority voting, while confidence-based weighting \cite{taubenfeld2025confidence} refines this selection. Tree of Thoughts (ToT) \cite{yao2024tree} enhances self-consistency by structuring reasoning as a tree search. Self-reflection \cite{renze2024self} allows LLMs to iteratively refine responses. However, LLMs remain prone to biases \cite{khatun2024study}, underscoring the need for multi-LLM systems to cross-verify answers and enhance reliability.

\section{Other Metrics}\label{appx:metric}

We compare our proposed uncertainty-aware metric with several commonly used token-level confidence and uncertainty metrics.

\paragraph{Entropy.}
Entropy captures the model's uncertainty over its output distribution:
\begin{equation}
\text{Entropy}(Y) = \frac{1}{T} \sum_{t=1}^T \sum_{v \in \mathcal{V}} -P(y_t{=}v \mid \cdot) \log P(y_t{=}v \mid \cdot),
\end{equation}
where \(\mathcal{V}\) denotes the vocabulary. Lower entropy indicates greater confidence in the model’s predictions.

\paragraph{Perplexity.}
Perplexity is the exponentiated negative average log-likelihood:
\begin{equation}
\text{Perplexity}(Y) = \exp\left(-\frac{1}{T} \sum_{t=1}^T \log P(y_t \mid \cdot)\right).
\end{equation}
Lower perplexity values indicate that the model considers the response more likely or fluent.

\paragraph{Gini Impurity.}
Gini impurity measures the dispersion of the output distribution:
\begin{equation}
\text{Gini}(Y) = \frac{1}{T} \sum_{t=1}^T \left(1 - \sum_{v \in \mathcal{V}} P(y_t{=}v \mid \cdot)^2 \right).
\end{equation}
A lower Gini score corresponds to a more confident (i.e., peaked) distribution over tokens.

\paragraph{KL Divergence (Model Disagreement).}
To quantify disagreement between models, we compute the average pairwise KL divergence:
\begin{equation}
\scalebox{0.85}{$
\text{KL}(\pi_i \parallel \pi_j) 
= \frac{1}{T} \sum_{t=1}^T \sum_{v \in \mathcal{V}} 
   P_{\pi_i}(y_t{=}v) 
   \log \frac{P_{\pi_i}(y_t{=}v)}{P_{\pi_j}(y_t{=}v)}
$}
\end{equation}
This can be averaged over all model pairs \((i, j)\) to measure the degree of inter-model uncertainty.

\section{Experimental Setup}\label{appx:experimental_setup}
\paragraph{Models Used:} We use the following model since they are all of same size and have quite comparable performance. However Qwen seems to be the strongest out of all. \\
1. Qwen2.5-7B-Instruct \\
2. Ministral-8B-Instruct-2410 \\
3. Llama-3.1-8B-Instruct.

\paragraph{Datasets Used:} We use the following datasets. 
\begin{enumerate}
    \item GSM8k \cite{cobbe2021training} is a math reasoning task based dataset. 
    \item MMLU \cite{hendrycks2020measuring} is a massive multitask test consisting of multiple-choice questions from various branches of knowledge. The test spans subjects in the humanities, social sciences, hard sciences, and other areas. We choose 8 subsets based on their closeness to reasoning task: 
        \begin{enumerate}
            \item formal-logic (FL)
            \item high-school-mathematics (HSM)
            \item elementary-mathematics (EM)
            \item college-mathematics (CM)
            \item high-school-computer-science (HSCS), \item philosophy (PHI)
            \item abstract-algebra (AA)
            \item high-school-statistics (HSS).
        \end{enumerate}
    \item ARC \cite{allenai:arc} is a grade-school level, multiple-choice science questions, assembled to encourage research in advanced question-answering. We choose the ARC-Challenge subset of this dataset. 
\end{enumerate}

\section{Intuitive Justification for our Metric}\label{appx:theory}

Below we show why our \emph{calibrated log‐likelihood} score
\[
\mathrm{Score}(Y)
\;=\;
\frac{1}{N}\sum_{i=1}^N
\frac{1}{T}\sum_{t=1}^T
\log P_{\pi_i}\bigl(y_t \mid x, X, y_{<t}\bigr)
\]
is fully comparable across different candidate answers \(Y\), and depends only on \(Y\) itself.

\paragraph{1. Per‐token, per‐model normalization}
Each model \(\pi_i\) assigns a joint probability to the sequence \(Y = (y_1,\dots,y_T)\):
\[
P_{\pi_i}(Y\mid x,X)
=\prod_{t=1}^T P_{\pi_i}\bigl(y_t \mid x, X, y_{<t}\bigr).
\]
Taking the logarithm and dividing by sequence length \(T\) yields the average per‐token log‐likelihood:
\[
\frac{1}{T}\sum_{t=1}^T
\log P_{\pi_i}\bigl(y_t \mid x, X, y_{<t}\bigr).
\]
Dividing by \(T\) removes any bias toward shorter or longer sequences, placing all candidates on the same per‐token scale.

\paragraph{2. Cross‐model averaging}
We then average these normalized log‐likelihoods across the \(N\) models:
\[
\mathrm{Score}(Y)
=\frac{1}{N}\sum_{i=1}^N
\underbrace{\frac{1}{T}\sum_{t=1}^T
\log P_{\pi_i}\bigl(y_t \mid x, X, y_{<t}\bigr)}_{\displaystyle\text{model }i\text{ score}}.
\]
Equivalently,
\[
\mathrm{Score}(Y)
=\frac{1}{T}\sum_{t=1}^T
\underbrace{\frac{1}{N}\sum_{i=1}^N
\log P_{\pi_i}\bigl(y_t \mid x, X, y_{<t}\bigr)}_{\displaystyle\text{consensus token score}}.
\]
Noting that
\[
\exp\bigl(\mathrm{Score}(Y)\times T\bigr)
=\Bigl(\prod_{i=1}^N P_{\pi_i}(Y\mid x,X)\Bigr)^{1/N},
\]
the score corresponds to the geometric mean of the per‐model probabilities, ensuring that a high score requires \emph{all} models to find \(Y\) likely.

\paragraph{3. Fixed hyperparameters \(\Rightarrow\) only \(Y\) varies}
All inference hyperparameters (e.g., temperature, tokenization, number of rounds \(K\)), as well as the number weights and the sequence length \(T\), are held constant when computing \(\mathrm{Score}(Y)\). Hence, the only variable across different candidate answers is the token sequence \(Y\) itself, making scores directly comparable.

\paragraph{Interpretation as a Product‐of‐Experts}
Define the product‐of‐experts distribution
\[
P_{\mathrm{prod}}(Y)
\;\propto\;\prod_{i=1}^N P_{\pi_i}(Y\mid x,X).
\]
Then
\[
\begin{aligned}
\log P_{\mathrm{prod}}(Y)
  &= \sum_{i=1}^N \log P_{\pi_i}(Y\mid x,X) \\[0.5em]
\Longrightarrow\quad
\mathrm{Score}(Y)
  &= \frac{1}{N\,T}\,\log P_{\mathrm{prod}}(Y)\,.
\end{aligned}
\]
Maximizing \(\mathrm{Score}(Y)\) is therefore equivalent to finding the answer \(Y\) that maximizes the product‐of‐experts likelihood, i.e.\ the response deemed most probable by the ensemble as a whole.

\begin{figure}[htbp]
  \centering
  \includegraphics[width=0.8\columnwidth]{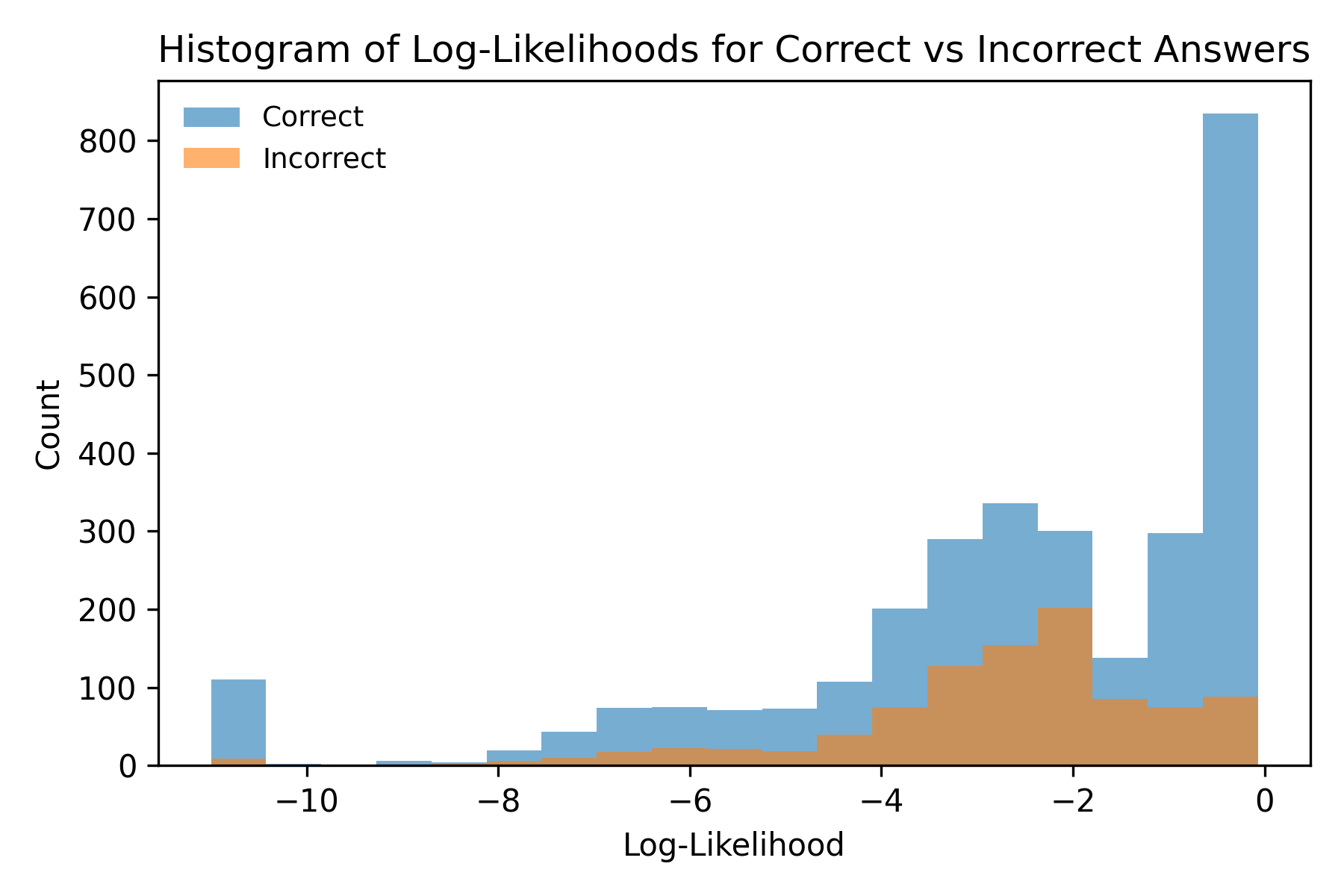}
  \caption{Histogram of log‐likelihood scores, showing the distribution for correct (blue) versus incorrect (orange) responses. The histogram is plotted over responses sampled from round 3 of the LLM-debate experiment from the GSM8K dataset.}
  \label{fig:ll_histogram}
\end{figure}

\paragraph{Conclusion}
By normalizing per token and averaging across a fixed set of models, all extraneous factors—sequence length, model‐specific scales, and hyperparameters—are eliminated. The resulting score depends solely on the candidate answer \(Y\), ensuring fair and robust comparison across multi‐LLM outputs.

\section{Additional Results}
\subsection{Comparison between applying calibrated log-likelihood metric for all cases vs tie-break cases only.}\label{appx:tie_break}

\begin{table}[htbp]
\centering
\resizebox{\columnwidth}{!}{%
\begin{tabular}{lccc}
\hline
\textbf{} & \textbf{Random} & \textbf{Tie-Break Case only} & \textbf{All cases} \\
\hline
GSM8K (Q=1, L=1, M=1)       & 47.87 & 70.00 & 71.00 \\
GSM8K Debate (Q, L, M)      & 81    & 84.88 & 85.03 \\
ARC (Q=1, L=1, M=1)         & 83.90 & 89.00 & 88.50 \\
\hline
\end{tabular}
}
\caption{Comparison of applying log-likelihood metric to tie-break only cases vs all cases.}
\label{tab:loglikelihood-results}
\end{table}

\end{document}